\documentclass[letterpaper, 10 pt, conference]{ieeeconf}  

\usepackage{times}
\usepackage{epsfig}
\usepackage{graphicx}
\usepackage{amssymb}
\usepackage{fontawesome}
\usepackage{threeparttable}
\usepackage{tikz}
\usepackage{comment}
\usepackage[utf8]{inputenc} 
\usepackage[T1]{fontenc}    
\usepackage{url}            
\usepackage{nicefrac}       
\usepackage{microtype}      
\usepackage[bb=boondox]{mathalfa}
\usepackage{amsmath}
\usepackage{multirow}
\usepackage{algpseudocode}
\usepackage{algorithm}
\usepackage{wrapfig}
\usepackage{rotating}
\usepackage{hyperref}
\usepackage{subcaption}
\usepackage{booktabs}       
\usepackage{amsfonts}       
\usepackage{marvosym}
\usepackage{pifont}
\usepackage{float}
\usepackage{caption}
\usepackage{bm}
\newcommand{\bo}{\mathbf{o}}
\newcommand{\bp}{\mathbf{p}}
\newcommand{\bv}{\mathbf{v}}

\usepackage[capitalize]{cleveref}
\crefname{section}{Sec.}{Secs.}
\Crefname{section}{Section}{Sections}
\Crefname{table}{Table}{Tables}
\crefname{table}{Tab.}{Tabs.}

\IEEEoverridecommandlockouts                              

\title{\LARGE \bf 
Multi-UAV Formation Control with Static and Dynamic  \\  Obstacle Avoidance via Reinforcement Learning
}

\author{Yuqing Xie$^{1*}$,
Chao Yu$^{1*}$\textsuperscript{\Letter},
Hongzhi Zang$^{1*}$, 
Feng Gao$^{1}$,
Wenhao Tang$^{1}$,\\
Jingyi Huang$^{+}$,
Jiayu Chen$^{1}$,
Botian Xu$^{+}$,
Yi Wu$^{1,2}$,
Yu Wang$^{1}$\textsuperscript{\Letter}
\thanks{* Equal Contributions.}
\thanks{{\Letter} Corresponding Authors. \tt{\{yuchao,yu-wang\}@tsinghua.edu.cn}}
\thanks{$^{+}$ Work done as an intern in Tsinghua University.}%
\thanks{$^{1}$ Tsinghua University, Beijing, 100084, China.}%
\thanks{$^{2}$ Shanghai Qizhi Institute.}%
\thanks{This research was supported by National Natural Science Foundation of China (No.62406159, 62325405), Tsinghua University Initiative Scientific Research Program, Tsinghua-Meituan Joint Institute for Digital Life, Beijing National Research Center for Information Science, Technology (BNRist) and Beijing Innovation Center for Future Chips, Postdoctoral Fellowship Program of CPSF under Grant Number GZC20240830, China Postdoctoral Science Special Foundation 2024T170496.}
}

\begin{document}
\newcommand{\xyq}[1]{\textcolor{red}{#1}}
\newcommand{\gf}[1]{\textcolor{orange}{[GF: #1]}}
\newcommand{\cmark}{\ding{51}}
\newcommand{\xmark}{\ding{55}}

\maketitle
\thispagestyle{empty}
\pagestyle{empty}

\setlength{\belowcaptionskip}{-10pt}
\setlength{\voffset}{10pt }
\setlength{\abovecaptionskip}{1pt} 

\begin{abstract}

This paper tackles the challenging task of maintaining formation among multiple unmanned aerial vehicles (UAVs) while avoiding both static and dynamic obstacles during directed flight. The complexity of the task arises from its multi-objective nature, the large exploration space, and the sim-to-real gap. To address these challenges, we propose a two-stage reinforcement learning (RL) pipeline. In the first stage, we randomly search for a reward function that balances key objectives: directed flight, obstacle avoidance, formation maintenance, and zero-shot policy deployment. The second stage applies this reward function to more complex scenarios and utilizes curriculum learning to accelerate policy training. Additionally, we incorporate an attention-based observation encoder to improve formation maintenance and adaptability to varying obstacle densities. Experimental results in both simulation and real-world environments demonstrate that our method outperforms both planning-based and RL-based baselines in terms of collision-free rates and formation maintenance across static, dynamic, and mixed obstacle scenarios. Ablation studies further confirm the effectiveness of our curriculum learning strategy and attention-based encoder. Animated demonstrations are available at: \url{https://sites.google.com/view/ uav-formation-with-avoidance/}.

\end{abstract}

\section{Introduction}
\label{sec:intro}

\begin{figure*}[h]
\centering
\includegraphics[width=1.0\textwidth]{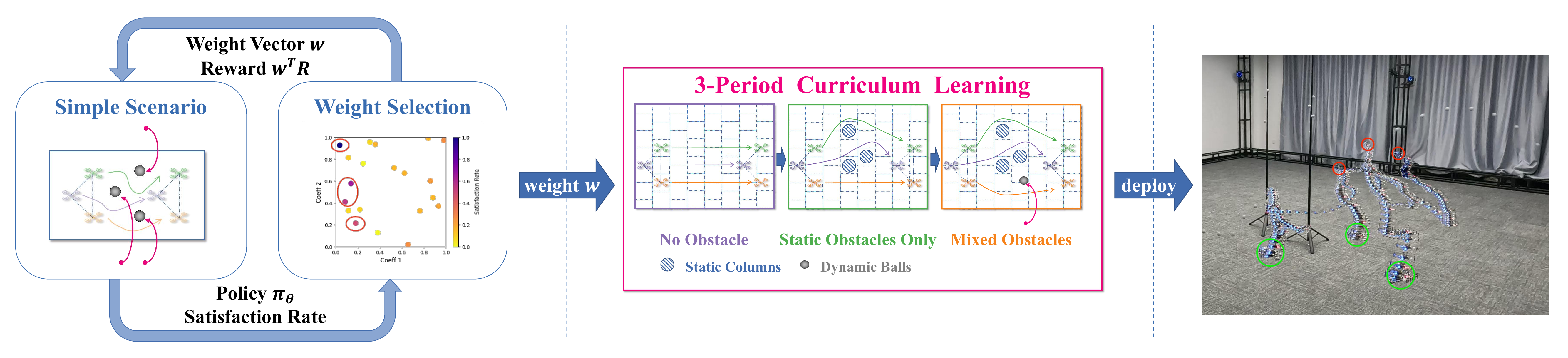}
\caption{Illustration of the proposed two-stage training pipeline. In the first stage, we randomly search within the reward function space to find the optimal function that balances all four objectives. In the second stage, we apply the selected reward function to a more complex scenario and solve the task with curriculum learning. }
\label{fig:overview}
\end{figure*}

The collaboration of multiple unmanned aerial vehicles (UAVs) has received vast attention due to its significant potential in practical applications, such as search and rescue missions \cite{liu2016multirobot} and payload transportation \cite{rao2023temporal}. However, this task is highly challenging, as it requires not only the basic capabilities of individual UAVs, such as flight control and obstacle avoidance, but also the effective coordination of all UAVs to achieve the collective objectives.

In this work, we focus on formation maintenance as the collective objective. Specifically, we address the challenge of multi-UAV formation control while ensuring both static and dynamic obstacle avoidance. 

Obstacle avoidance is crucial in multi-UAV collaboration. In addition to navigating around environmental obstacles, as in single-UAV scenarios, UAVs in a multi-agent system must also avoid collisions with one another. Traditional planning-based methods \cite{swarm1, swarm2, edg, 30uav, pso} generate collision-free paths for UAVs to handle static obstacles. Dynamic obstacles, which are common in real-world environments, introduce additional complexity. Predictive planning for dynamic obstacles \cite{mader, kondo2023robust, puma, dream, mrnav} often requires prior knowledge of obstacle trajectories, limiting its applicability. With the advancement of deep reinforcement learning (RL), learning-based methods \cite{batra2022decentralized, han2019multi} have shown promising results in enabling multi-UAV collision avoidance in environments with both static and dynamic obstacles.

Formation control, the collective objective of our task, is also a critical aspect of multi-robot collaboration. However, when tackling formation control, the obstacle avoidance requirement is often simplified or overlooked. Planning-based methods \cite{formation1, quan2023robust} typically focus on scenarios without obstacles or with only static obstacles. Similarly, learning-based approaches \cite{f-rl-1, f-rl-2, f-rl-4} utilize multi-agent reinforcement learning (MARL) for multi-UAV formation control in obstacle-free environments, and \cite{f-rl-5, f-rl-6} apply MARL to ground vehicle formation control with static obstacle avoidance only.

To the best of our knowledge, our work is the first to tackle the combined challenge of maintaining formation while avoiding both static and dynamic obstacles. This task is particularly challenging due to its multi-objective nature, where the UAVs must balance several potentially conflicting goals, such as directed flight, formation maintenance, and obstacle avoidance. Additionally, compared to ground vehicles, UAVs require their RL policy to tackle a larger sim-to-real gap and a broader exploration space, as they navigate in 3D space and have higher mobility.

To address these challenges, we propose a two-stage reinforcement learning (RL) training pipeline. In the first stage, we search for a reward function that balances four task objectives, including directed flight, obstacle avoidance, formation maintenance, and action smoothness. The former three completes the task, and the last one bridges the sim-to-real gap. We evaluate the performance of different reward functions in a simpler scenario and select the one that best aligns with our preferences. In the second stage, we apply the selected reward function to tackle a more complex task setting. To accelerate training and enhance final performance, we employ curriculum learning (CL), which progressively increases task difficulty. Additionally, we design an attention-based observation encoder to handle varying numbers of obstacles and to improve formation maintenance among multiple UAVs.

Our main contributions are as follows:
\begin{itemize}
\item We design a two-stage training pipeline to address the multi-objective task of formation maintenance and obstacle avoidance for both static and dynamic obstacles.
\item We conduct extensive simulations to demonstrate the effectiveness of our proposed training pipeline and network architecture. Our method outperforms both planning-based and RL-based baselines regarding collision-free rate and formation maintenance, and can easily generalize to varying number of obstacles. 
\item We deploy the learned policy in real-world environment without additional fine-tuning, showcasing its practical applicability and robustness.
\end{itemize}

\section{Related Work}

\subsection{Multi-UAV Formation}

Extensive research has been conducted on multi-UAV formation.
Classical planning-based methods \cite{formation1, formation2} and emerging learning-based methods \cite{f-rl-1, f-rl-2, f-rl-4, f-rl-3} guide UAVs to form a desired configuration from random initial positions. Compared to the leader-follower schemes \cite{f3, f4}, fully decentralized methods \cite{swarm-formation, f6, f7, f8} offer better scalability and resiliency to partial failures. However, these algorithms address only obstacle-free or static-obstacle scenarios and do not account for dynamic obstacles, such as flying birds. Additionally, some RL-based methods focus on navigating through static obstacles while maintaining formation in ground vehicle systems \cite{f-rl-5, f-rl-6}. In this paper, we derive an RL policy for UAV formation control in a fully decentralized manner. The transition from 2D to 3D space significantly enlarges the exploration space for RL algorithms, while incorporating dynamic obstacles introduces greater challenges for obstacle avoidance strategies.

\subsection{Collision Avoidance for UAV}
Obstacle avoidance is crucial for UAVs to ensure safe flights in cluttered environments. Methods like PANTHER \cite{panther} and Deep-PANTHER \cite{deeppanther} effectively navigate a single UAV through obstacle-rich environments. In multi-UAV scenarios, UAVs must avoid not only environmental obstacles but also potential collisions with other UAVs. Planning-based approaches leverage gradient-based optimization \cite{swarm1, swarm2, edg}, evolutionary optimization \cite{30uav}, or particle swarm optimization \cite{pso} to plan collision-free paths for multiple UAVs in constrained environments, such as forests. However, these methods are designed for static environments, while real-world scenarios involve dynamic obstacles.  Avoiding dynamic obstacles is particularly challenging, as UAVs must respond quickly to sudden obstacles and execute aggressive maneuvers \cite{obs2, obs3}. Predictive planning methods usually require additional knowledge of obstacle trajectories, either provided \cite{mader, kondo2023robust} or predicted \cite{puma, dream, mrnav}. Although planning-based approaches guarantee collision-free trajectories and can operate in real-time, they constrain UAV behavior and face a trade-off: decentralized planning may lead to deadlocks, while centralized planning reduces adaptability.

RL shows promise in overcoming these limitations for multi-UAV collision avoidance. \cite{batra2022decentralized} trains an RL policy to control motor thrust for multi-UAV tracking in obstacle-free environments, which can be adapted to collision avoidance by modeling obstacles as UAVs. Similarly, \cite{han2019multi} treats multi-UAV collision avoidance as a single UAV obstacle avoidance problem. In this paper, we employ RL to tackle the complex task of multi-UAV collision avoidance in mixed obstacle scenarios while maintaining formation. We determine the optimal reward function for multiple objectives in simpler scenarios and then transfer this function to more challenging settings, thereby balancing computational requirements and policy efficacy.

\section{Task}

\subsection{Task Setup}
\label{sec:task}
We address the task of multi-UAV formation maintenance with obstacle avoidance during directed flight. The task aims to achieve four objectives:
\begin{itemize}
  \item \textbf{Directed Flight.} The drones should fly together in a specified direction, following a target velocity. 
  \item \textbf{Formation Maintenance.}  The drones should maintain a predefined formation shape and size but can rotate the arrangement. For agile obstacle avoidance, the drones will not receive manually assigned target positions; instead, they must determine their own positions based on the formation centroid and the status of other drones.
  \item \textbf{Obstacle Avoidance.} The drones must avoid all environment obstacles, both dynamic and static, and prevent collisions with each other.
  \item \textbf{Action Smoothing.} To facilitate zero-shot sim-to-real transfer and reduce energy consumption, the policy should generate smooth and consistent action commands.
\end{itemize}
We initialize all drones at their respective positions within the desired formation. Multiple obstacles are placed along their flight path. We partition the environment using a zigzag grid arrangement, where each grid cell measures 0.5m with a 0.25m horizontal offset between adjacent rows, as shown in Fig. \ref{fig:overview}. 

From these grids, we randomly sample multiple cells to place static obstacles, which are columns with a radius of 0.15m and height of 3m. Dynamic obstacles are balls with a radius of 0.15m, moving in a parabolic trajectory towards a randomly chosen drone.

\subsection{Observation Space}
The observation $\bo^i$ for drone $i$ is composed of three components: self-information, $\bo_{self}$; information about other drones, $\bo_{drones}$; and details regarding static and dynamic obstacles, $\bo_{static}$ and $\bo_{dynamic}$.

$\bo_{self}$ contains a 3D position coordinate $\mathbf{p}$, a 4D quaternion indicating its orientation $\mathbf{q}$, a 3D linear velocity $\bv$, and a 3D angular velocity $\bm{\omega}$. To better capture the drone's orientation, we rotate the x-axis and z-axis of the world coordinate using $\mathbf{q}$, producing 3D vectors $\mathbf{h}$ and $\mathbf{u}$. $\bo_{self}$ also includes the drone's relative velocity $\bv_{rel}$ compared to the reference velocity $\bv_{target}$ and a 3D representation of the drone's identifier.
$\bo_{drones}$ covers the L2 distance, relative position, and relative velocity of other drones relative to drone $i$. 

Following \cite{huang2023collision}, we calculate the distance from $3 \times 3$ grid cells around the drone to the nearest obstacles and use the $3 \times 3$ distance tensor to characterize all obstacles. Unlike raw positional observations, this quantity- and permutation-invariant representation enhances the policy’s understanding of spatial data. We adopt this method and acquire a 9-dimensional $\bo_{static}$ for each drone. $\bo_{dynamic}$ includes the L2 distance, relative position, relative velocity, and absolute velocity of the obstacle.

\subsection{Action Space}
Inspired by \cite{kaufmann2022benchmark}, we adopt collective thrust and body rates (CTBR) commands as policy actions to ensure robust sim-to-real transfer and agile control. These CTBR commands are then converted to motor thrust commands using PID controllers. The action for drone $i$ is expressed as $\mathbf{a}=(c, \omega_{roll}, \omega_{pitch}, \omega_{yaw})$, where $c \in [0, 0.9]*$max\_thrust indicates the  collective thrust, and $\omega \in [-\pi, \pi]$ signifies the body rates for the corresponding axes.

\section{Preliminary}

Our task inherently involves multi-agent cooperation and multi-objective optimization. Therefore, to model our task, we extend the definition of Markov decision processes as multi-objective, decentralized, partially observable Markov decision processes (MO-Dec-POMDPs). 
A MO-Dec-POMDP is defined as $<S, A, O, \textbf{R}, P, n, \gamma>$, where
$S: \mathbb{R}^{d_{s}} $ is the state space,
$A: \mathbb{R}^{d_{a} \times n}$ is the joint action space,
$O: \mathbb{R}^{d_{obs} \times n}$ is the partial observation space,
$\textbf{R}: S\times A \rightarrow \mathbb{R}^{d_{obj}}$ is a vector-valued reward function that generates the reward for each objective,
$P: S\times A \times S\rightarrow [0,1]$ is the transition probability from state $s_t$ to $s_{t+1}$ when taking joint action $a_t$,
$n$ is the number of agents, and
$\gamma$ is the discount factor for rewards. $d_s, d_a, d_{obs}, d_{obj}$ represent the dimensions of the state, action, observation, and objective, respectively.

Solving the above MDP takes two steps. Firstly, we transform the multi-objective setting to a single-objective setting through a utility function $u: \mathbb{R}^{d_{obj}} \rightarrow \mathbb{R}$ and thus obtain a scalar reward $R$ function, $R: S\times A \rightarrow \mathbb{R}$. 
Then, as the above MO-Dec-POMDP becomes a standard Dec-POMDP, we apply multi-agent reinforcement learning algorithms to find an approximate solution. 
\section{Methodology}
We introduce a two-stage RL training pipeline to solve our proposed multi-objective task, as illustrated in Fig.~\ref{fig:overview}. In the first stage, we identify a linear utility function $ u(\textbf{R})=\textbf{w}^\text{T}\textbf{R}$ that properly balances between all objectives in a simplified scenario. 
Once the weight vector \textbf{w} is determined, we transition to the second stage, where we use the utility function to tackle more complex task scenarios. We use curriculum learning (CL) to navigate large exploration spaces and accelerate training.

In this section, we detail the multi-objective reward function and our two-stage training pipeline. We also describe the network architecture, which features an attention-based observation encoder to improve formation control and deal with different quantities of obstacles.

\subsection{Multi-Objective Reward Function}
\label{sec:reward}

As outlined in Sec. \ref{sec:task}, the task encompasses four objectives: directed flight, formation maintenance, obstacle avoidance, and action smoothing. Each objective is associated with a scalar reward component. 
Prior to detailing each reward component, we define three functions to characterize the relationship between reward $r$ and error $e$. 
\begin{itemize}
    \item Linear: $r_l(e)=1-|e|$.
    \item Reciprocal: $r_r(e)=1/(1+e)$.
    \item Indicator: $r_i(e)=\mathbb{1}_{e>e_{thres}}$.
\end{itemize}
The first two functions ensure that when $e$ ranges from 0 to 1, $r$ remains within the same interval, and $r$ ascends as $e$ declines. The last function imposes a penalty when the error surpasses a specified threshold.

\textbf{Flight Reward} $R_{flight}$ involves constraints related to altitude error $e_{height}$, reference velocity error $e_{v}$, reference position error $e_{p}$, and heading error $e_{heading}$. The flight reward is expressed as $R_{flight}=\alpha_{heading}*r_l(e_{heading})+\alpha_{v}*r_l(e_{v})+\alpha_{p}*r_r(e_{p})+\alpha_{height}*r_l(e_{height})$. 

\textbf{Formation Reward} $R_{formation}$ includes constraints related to formation size, shape, and the distances between drones. Following \cite{swarm-formation}, we use the Laplacian distance $e_{shape}$ to measure formation similarity between the target formation and the current formation. 
We measure the pairwise distances between all drones, denoting the smallest as $dis$ and the largest as $size$. The size error is $e_{size}=(size-size_{target})^2$. Thus, we obtain $R_{formation}=\alpha_{shape}*r_r(e_{shape})+\alpha_{size}*r_r(e_{size})+\alpha_{dis}*r_i(dis)$.

\textbf{Obstacle Reward} $R_{obstacle}$ is determined by the distance from the drone to the closest obstacle. This reward only appears when obstacles are within the drone's observation range. 

\textbf{Action Reward} $R_{action}$ motivates the policy to generate smooth actions, quantified by the per-rotor difference in throttle between successive time steps. We evaluate the network output $e_{net}=\|\mathbf{a}_t-\mathbf{a}_{t-1}\|_2$, the drone's throttle difference $e_{diff}=\|\mathbf{throttle}_t-\mathbf{throttle}_{t-1}\|_2$, the total throttle of all 4 rotors $e_{throt}=\sum \mathbf{throttle}_t$, and the rotation along the yaw axis $e_{yaw}$. Thus, the action reward is $R_{action}=\alpha_{net}*r_l(e_{net})+\alpha_{diff}*r_l(e_{diff})+\alpha_{throt}*r_l(e_{throt})+\alpha_{yaw}*r_l(e_{yaw})$.

Note that the $\alpha$ coefficients are set empirically to ensure that each term within the reward component is on a similar scale.

\subsection{Two-Stage Training Pipeline}

\subsubsection{The First Stage, Reward Scalarization}

In this stage, we transform the multi-objective vector reward $\textbf{R} = (R_{formation}, R_{flight}, R_{obstacle}, R_{action})\in\mathbb{R}^{d_{obj}}$ into a scalar reward $R$ automatically. We assume $R$ is a weighted sum of the reward components, $R = u(\textbf{R}) = \textbf{w}^{\text{T}}\textbf{R}$, with weight vector $\textbf{w} = (w_{formation}, w_{flight}, w_{obstacle}, w_{action})$, $\|\mathbf{w}\|_1=1 $.
The goal is thus to search for a set of $\textbf{w}$ that best aligns with our preference.

Existing MORL methods are not suitable for this problem as they cannot fully explore the weight space when the value of each objective becomes highly non-monotonic with the weight. Thus, to scalarize the reward vector, we repeatedly sample a random $\textbf{w}$ and use MAPPO \cite{yu2103surprising} to solve the transformed Dec-POMDP. 
For each $\textbf{w}$, we evaluate its corresponding RL policy using the satisfaction rate (SR) as the metric. An episode is considered satisfying if all four objectives meet predefined thresholds, and SR is the proportion of satisfying episodes out of the total episodes. In practice, we select the weight vector with the highest SR for the second stage.

However, finding the appropriate weight vector requires multiple trials, which is computationally expensive and time-consuming. To accelerate the convergence of MAPPO in each trial, we simplify the task to three drones maintaining a triangular formation while avoiding three dynamic obstacles.

\subsubsection{The Second Stage, Reward Generalization}

With a $\textbf{w}$ that properly scalarizes $\textbf{R}$, we can now tackle a more complex task: three drones navigating through both static and dynamic obstacles. This stage demonstrates the generalizability of the derived reward function and highlights the effectiveness of our proposed approach.

Training the RL policy from scratch for complex tasks is computationally intensive and often leads to suboptimal solutions. Hence, we utilize curriculum learning to speed up training through gradually increasing task difficulty. Specifically, we implement a 3-period curriculum, where drones first learn to fly forward in an obstacle-free environment, followed by a static-obstacle-only environment, and ultimately a mixed-obstacle environment.





%

\subsection{Network Architecture}

\subsubsection{Attention-Based Observation Encoder}
\begin{figure}[h]
\centering
\includegraphics[width=0.5\textwidth]{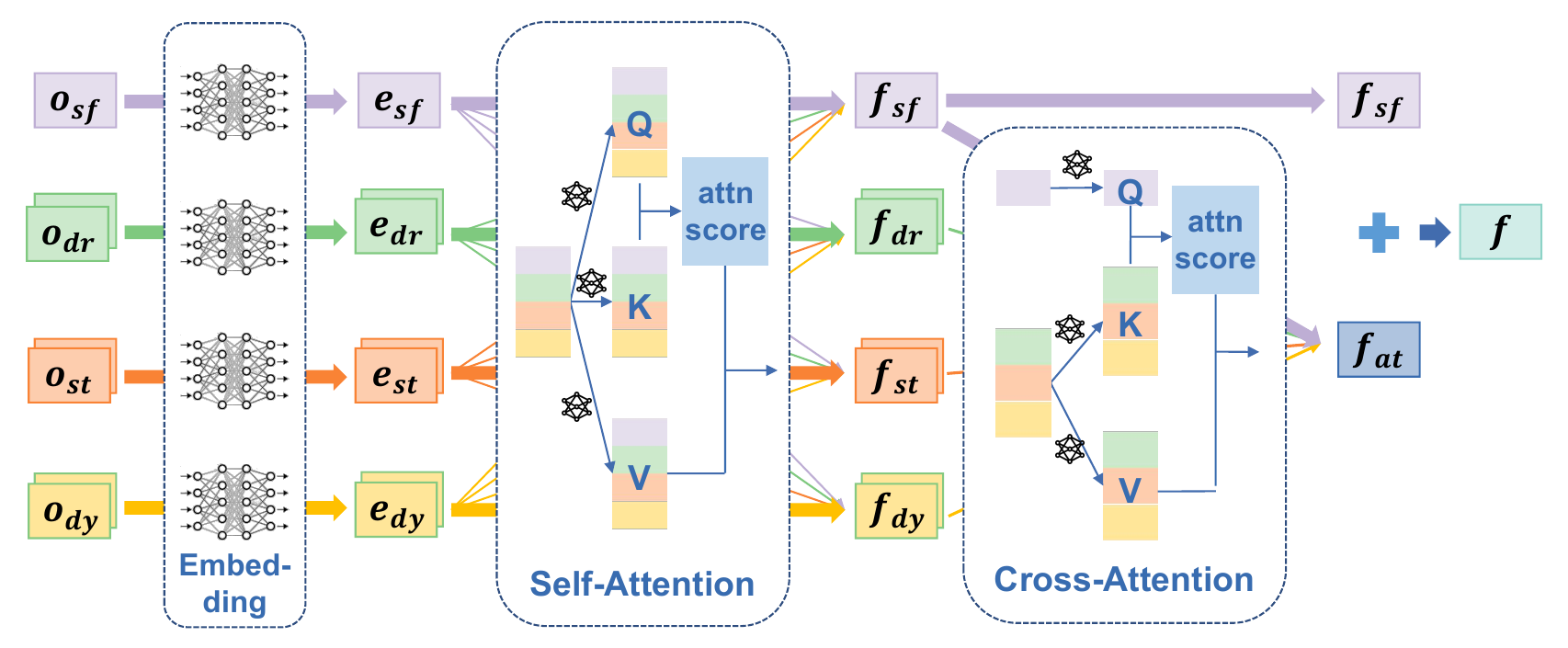}
\caption{The network structure of the observation encoder. The subscripts \textit{sf}, \textit{dr}, \textit{st}, and \textit{dy} are abbreviations for \textit{self}, \textit{drones}, \textit{static}, and \textit{dynamic}, as mentioned in the main text.}
\label{fig:encoder}
\end{figure}

To efficiently manage scenarios with different quantities of obstacles, we develop an attention-based observation encoder. This encoder eliminates the limitations on input dimensionality and emphasizes relevant task information. The structure of the encoder is shown in Fig.~\ref{fig:encoder}. The observation composes of four parts: $\bo_{self}$, $\bo_{drones}$, $\bo_{static}$, and $\bo_{dynamic}$. Each part is individually encoded through separate MLPs to produce embeddings of identical dimensions. We use multi-head self-attention on these embeddings to obtain respective features. To better capture the relationships between the drone and other environmental entities, these features are processed through a multi-head cross-attention module, with self-feature as the query and other features as the key and value. The final feature $\mathbf{f}$ is a concatenation of the self-feature and the output of the multi-head cross-attention.



\subsubsection{Actor\&Critic Network}

Following the observation encoder, we use a Gaussian distribution model $\pi_\theta = \mathcal{N}(\bm\mu_{\theta}, \bm\sigma_{\theta})$ to parameterize actions in the actor network, and an MLP $V_\phi(\mathbf{f})$ to estimate the value for the current state  in the critic network.

\
\section{Experiments}



We compare our approach with three state-of-the-art baselines that meet some objectives: Swarm-Formation \cite{quan2023robust}, R-Mader \cite{kondo2023robust}, and Swarm-RL \cite{huang2023collision}. The first two are planning-based algorithms, and the third one is RL-based algorithm.
Since all baselines fail to meet all four objectives in our task, we make the following adaptations:
\begin{itemize}
    \item \textbf{Swarm-Formation \cite{quan2023robust}} is designed to handle formation maintenance around static obstacles. We integrate a trajectory prediction module to facilitate dynamic obstacle avoidance. Upon detecting a dynamic obstacle, the module captures two sequential frames of the obstacle and fits a parabolic curve to estimate its trajectory. This estimated parabolic curve is then treated as a static obstacle and sent to the planner.
    \item\textbf{R-Mader \cite{kondo2023robust}} focuses on avoiding both static and dynamic obstacles for multiple drones without considering formation. To incorporate formation, we set the desired formation as the target positions of the drones. For dynamic obstacle avoidance, R-Mader requires knowledge of the obstacles' trajectories. Therefore, we fit parabolic curves to the obstacles and provide the curve parameters to R-Mader. 
    \item \textbf{Swarm-RL \cite{huang2023collision}} focuses on collision avoidance for static obstacles and inter-drone collisions but does not handle dynamic obstacles or formation maintenance. To address this, we incorporate the relative position and velocity of dynamic obstacles into its observation. Additionally, we introduce a velocity tracking reward to aid in training. As with R-Mader, we set the desired formation as the final positions of each drone. 
    
\end{itemize}

\subsection{Simulation Results}

We choose OmniDrones \cite{xu2024omnidrones} as our simulator because of its high-speed GPU simulation and realistic environment dynamics for drones.

We consider a challenging task: 3 drones fly as a near-equilateral triangle while avoiding static and dynamic obstacles scattered within the space. 
We obtain the policy through our proposed 2-stage RL training pipeline. 
During the first training stage, each trial requires 100 million environment steps. 
During the second training stage, we perform 3-period curriculum learning, where we train the RL policy in scenarios with no obstacle, 10 static obstacles only, and mixed obstacles with 10 static and 2 dynamic obstacles, sequentially. Each period takes 15 million, 150 million, and 150 million environment steps, respectively.
Swarm-RL takes 1 billion environment steps to converge, about 3x of our method. 

\subsubsection{Evaluation Metrics}

We consider two metrics to examine the performance:
\begin{itemize}
    \item \textbf{Collision-Free Rate (CFR).} 
    An episode is considered successful if drones avoid collisions with all obstacles and with each other while reaching a specified area in time. 
    The collision-free rate is defined as the proportion of successful episodes out of the total episodes. This metric evaluates the performance of directed flight and obstacle avoidance. The higher, the better.
    \item \textbf{Formation Maintenance (FM).} In successful episodes, we compute the unnormalized Laplacian distance $e_{shape}$ between the target formation and the actual swarm configuration averaged over the episode. Unlike the normalized metric used during training, this measurement is rotation-invariant but sensitive to size. Hence, a smaller unnormalized Laplacian distance indicates a more desirable formation in terms of both shape and size.
    
    

\end{itemize}

Note that we do not quantify the performance of action smoothing, since this term facilitates sim2real transfer, which will be verified in real-world experiments. 
In the following section, we report the CFR and FM of planning-based methods averaged over 25 different trials, as they do not support parallel evaluation, and RL-based methods over 100 trials. The positions of all obstacles are randomly generated in each trial.


\subsubsection{Main Results}\label{sec:main-results}
\begin{figure}[H]
    \centering
    {\includegraphics[width=0.98\linewidth]{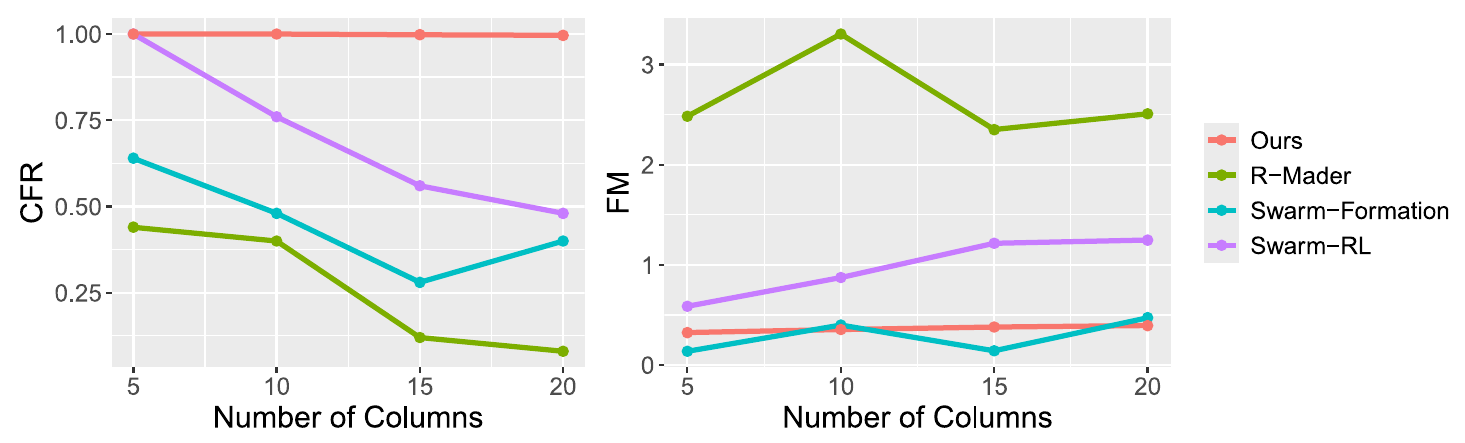}}
    \caption{Varying the number of static obstacles.}
    \label{fig:col_num}
\end{figure}

\begin{figure}[H]
    \centering
    {\includegraphics[width=0.98\linewidth]{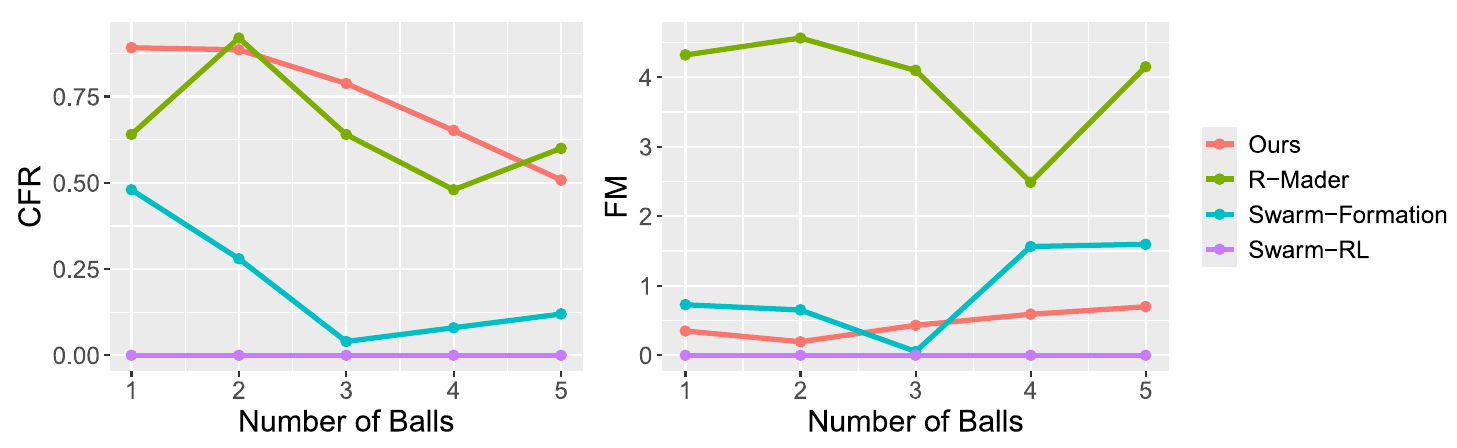}}
    \caption{Varying the number of dynamic obstacles.}
    \label{fig:ball_num}
\end{figure}

We test all methods in the mixed obstacle scenario with 2 balls and 10 columns and report the performance in Tab. \ref{table:simulation}. 
We further test the zero-shot transfer ability of all methods by varying the number of obstacles. We gradually increase the number of columns from 5 to 20 under static obstacle scenario, and the number of balls from 1 to 5 under dynamic obstacle scenario. The results are shown in Fig. \ref{fig:col_num} and Fig. \ref{fig:ball_num}, respectively. 

\begin{table}[H]
\centering
\begin{tabular}{ccccc}
 &  Swarm-Formation & R-Mader & Swarm-RL & Ours \\
  \hline
 \hline
 CFR($\uparrow$) & 0.08 & 0.20 & 0.83 & \textbf{0.89} \\
 FM($\downarrow$) &  1.353 &  4.052 & 0.959 & \textbf{0.278} \\
\end{tabular}
\caption{Performance of all methods under mixed obstacle scenario (2 balls + 10 columns) in simulation.}
\label{table:simulation}
\end{table}

\setlength{\belowcaptionskip}{1pt} 

\begin{figure*}
\centering

\subcaptionbox{Static obstacle scenario with 4 columns. The black boxes represent static obstacles. }{\includegraphics[width=\textwidth]{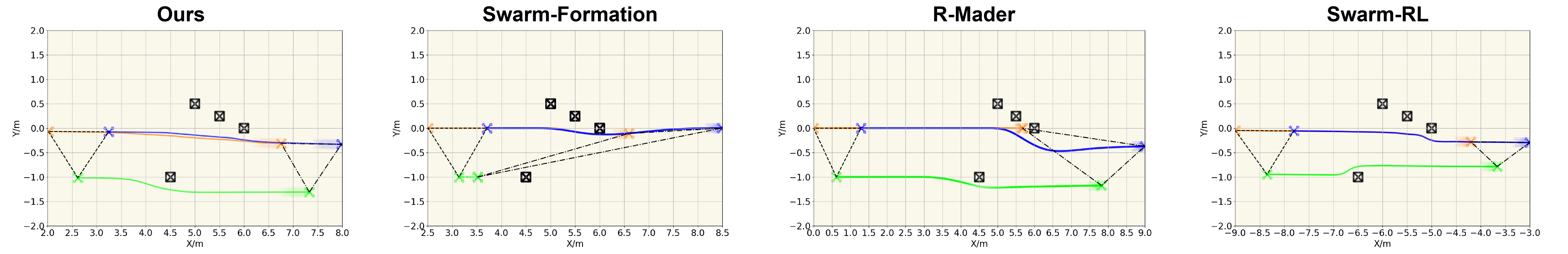}}
\subcaptionbox{Dynamic obstacle scenario with 1 ball. The red circles represent the trajectory of dynamic obstacles. }{\includegraphics[width=\textwidth]{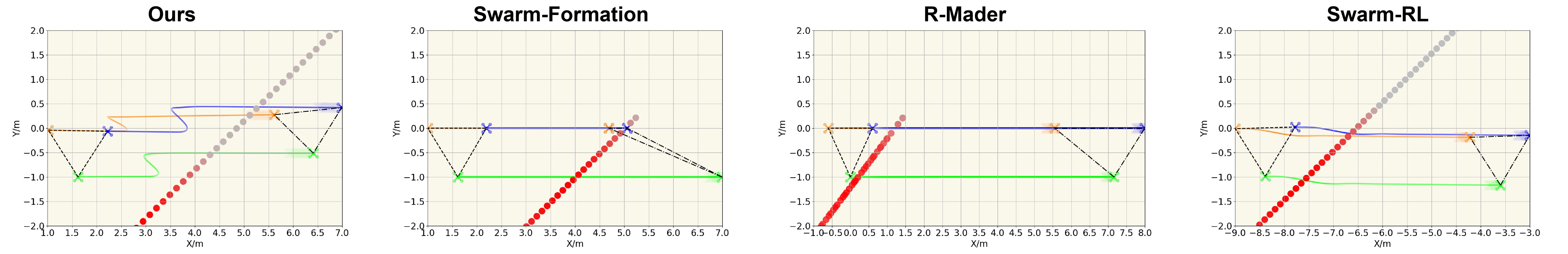}}
\caption{Behavior visualization.}
\label{fig:behavior}
\end{figure*}

\begin{figure*}
\centering
    \subcaptionbox{Satisfaction Rates}
    {\includegraphics[width=0.28\textwidth]{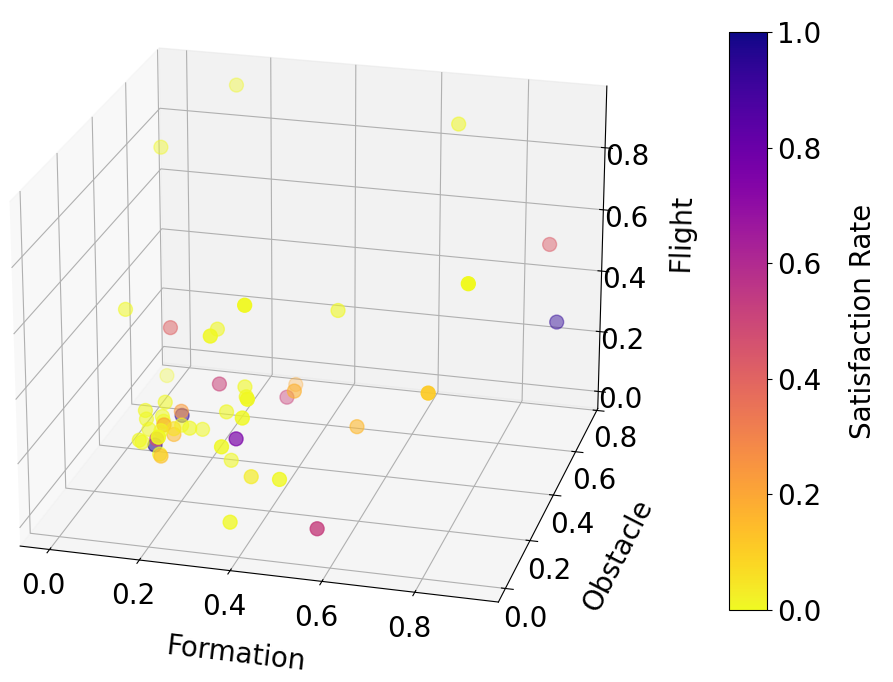}}
    \subcaptionbox{Diverse Behaviors}
    {\includegraphics[width=0.7\textwidth]{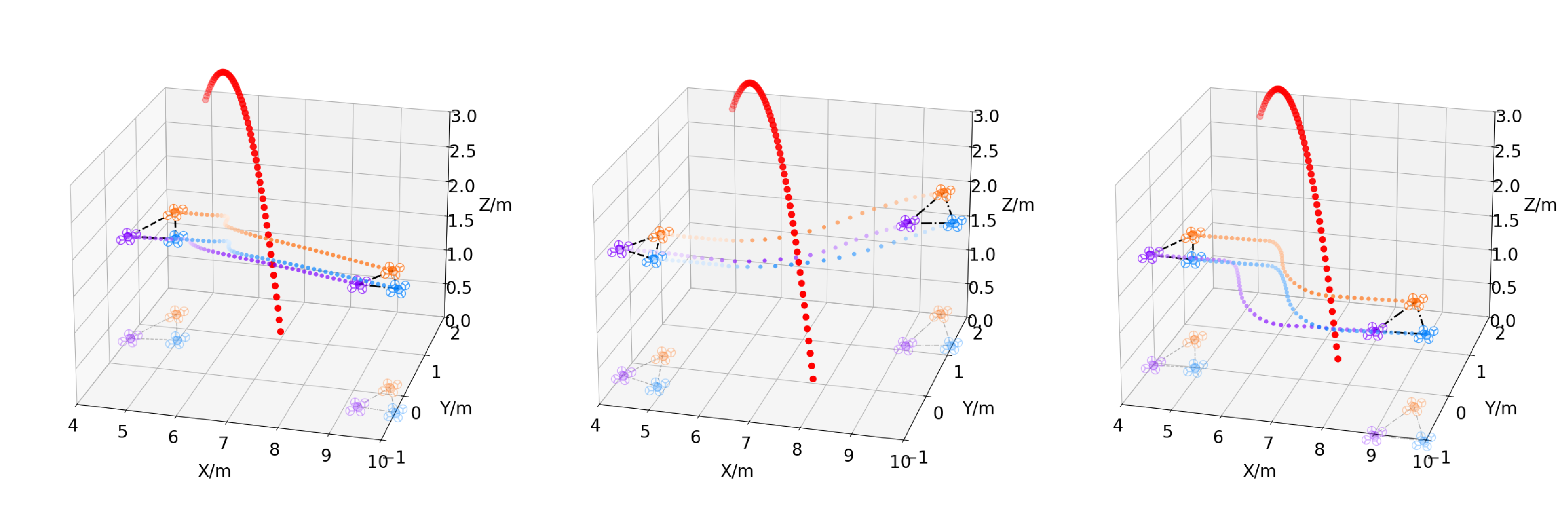}}  
    \caption{(a) The satisfaction rates under different weight vectors. The x-, y-, and z-axis represent the weight of the formation, obstacle, and flight objectives, respectively, and the weight of the action objective can be calculated as $w=1-x-y-z$. The darker the color, the higher the satisfaction rate. (b) Policy behaviors induced by different weight vectors. The red line represents the trajectory of the dynamic obstacle.}
    \label{fig:abl_weight}
\end{figure*}



Our method outperforms all baseline methods regarding both CFR and FM in the mixed obstacle scenario, and demonstrates comparable, if not superior, performance to the best baseline in extremely confined environments with 20 columns or 5 balls. The results demonstrate our method's strong capability to maintain effective formation while evading various obstacles, underscoring its ability to accomplish multi-objective tasks. Also, as the number of obstacles increases, our method shows only a slight degradation in both metrics, demonstrating its strong scalability and generalization capability, which we attribute to the attention mechanism employed in the network structure.

Swarm-Formation has a low CFR because it is designed for large-scale environments but not for fine-grained, dense obstacle distributions, as in our setting. It also lacks proper trajectory prediction for dynamic obstacles. Therefore, when a dynamic obstacle suddenly appears in front of a drone, the algorithm prematurely assumes a collision and ceases to control the drone.

R-Mader, which inherently incorporates dynamic obstacle avoidance, performs better than Swarm-Formation in dynamic scenarios but still falls short of our method. Like Swarm-Formation, R-Mader struggles to navigate through dense distributions of fine-grained obstacles, leading to a low CFR in static scenarios. Moreover, R-Mader is not designed for formation maintenance. As a result, the close proximity of drones in a dense formation makes it difficult to find a collision-free solution.


While Swarm-RL maintains high CFR in static obstacle scenarios, its formation is easy to break: when one drone changes its behavior to avoid obstacles, other drones simply ignore it and keeps flying forward. 
In dynamic settings, as the number of obstacles increases, the CFR drops significantly, indicating inadequate scalability.

\subsubsection{Behavior Visualization}

In Fig. \ref{fig:behavior}(a), we visualize the behaviors of different methods in the same scenario with 4 columns. Our method achieves obstacle avoidance by adjusting the centroid of the formation and slightly twisting its shape, demonstrating its flexibility in formation maintenance. Swarm-Formation and R-Mader fail to find paths around the fine-grained obstacles. Therefore, the green drone in Swarm-Formation and the orange drone in R-Mader remain stationary when other drones pass through. Swarm-RL fails to maintain formation during obstacle avoidance.

Fig. \ref{fig:behavior}(b) visualizes behaviors in scenario with 1 ball. 
Except for Swarm-Formation, all methods successfully avoid the ball.
Our method maintains formation by having all drones decelerate upon encountering the ball and then resume flight after it drops.
In R-Mader, the orange drone decelerates and falls behind the other drones, thereby disrupting the formation.
In Swarm-RL, the orange drone circles around the ball and rejoins the other drones after avoiding the obstacle.







\subsection{Analysis of Reward Scalarization}
\label{sec:metric}

Fig. \ref{fig:abl_weight} (a) illustrates the satisfaction rates of policies trained with randomly sampled weight vectors \textbf{w}. We observe that weight vectors with similar satisfaction rates can yield diverse behaviors. Fig. \ref{fig:abl_weight} (b) depicts the trajectories of 3 drones dodging a dynamic obstacle, guided by policies with similar satisfaction rates. When encountering a dynamic obstacle, the first policy and the third policy decelerate the drones in different magnitudes, whereas the second policy choose to let the drones ascend and accelerate. This suggests that our method holds strong flexibility and policy diversity, allowing humans to choose behaviors that best align with their preferences. In practice, we select the weight vector with the highest satisfaction rate.




\setlength{\textfloatsep}{1pt }
\subsection{Ablation Studies}

\begin{figure}
\centering
  \subcaptionbox{Observation Encoder}
    {\includegraphics[width=0.23\textwidth]{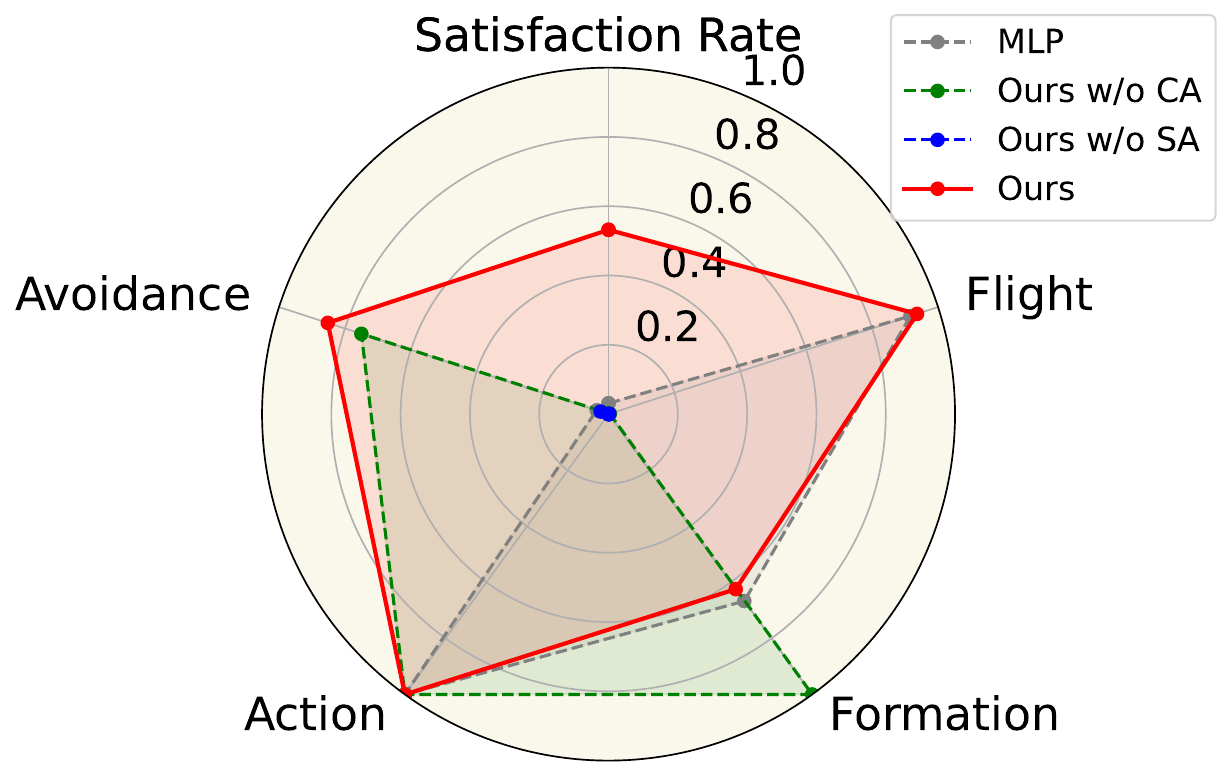}}
  \subcaptionbox{Curriculum Learning}
    {\includegraphics[width=0.23\textwidth]{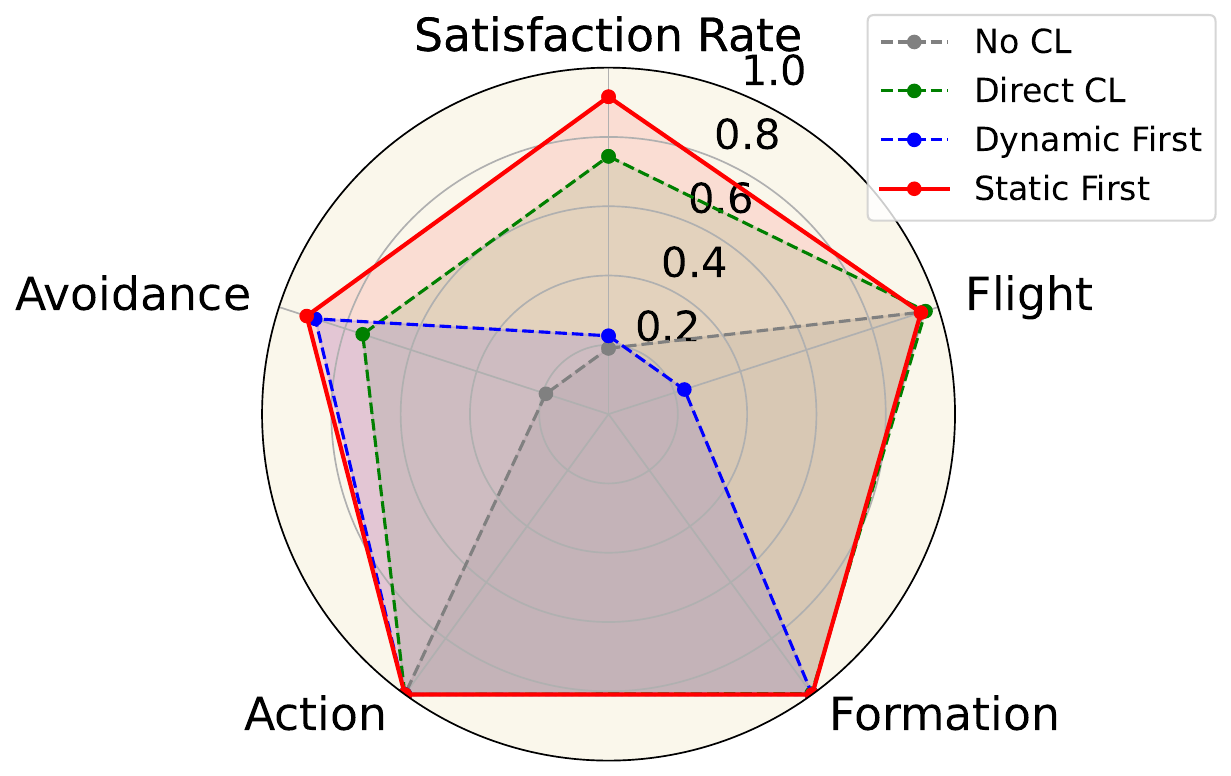}}
\caption{Results of the ablation studies.}
\label{fig:abs}
\end{figure}

In this section, we perform ablation studies on the attention-based observation encoder and curriculum learning used in the second stage. All experiments are conducted in a mixed obstacle scenario with 10 columns and 2 balls. For each objective, we define success criteria and record the proportion of episodes that meet these criteria. Additionally, we calculate the Satisfaction Rate, which represents the proportion of episodes that satisfy all criteria simultaneously.
%

\subsubsection{Attention-based Observation Encoder}

We compare three variants: (1) Ours w/o CA, where the cross-attention (CA) module is replaced with an MLP, (2) Ours w/o SA, where the self-attention (SA) module is replaced with an MLP, and (3) MLP, where both the cross-attention and self-attention modules are substituted with an MLP. 

As demonstrated in Fig. \ref{fig:abs}, only our method exhibits satisfying Success Rates regarding all objectives. The variants either fail to reach the destination or collide with obstacles. This suggests that the attention modules significantly enhance the policy’s ability to handle obstacles and integrate information from other drones.




\subsubsection{Curriculum Learning}

We consider the following four curriculum designs: (1) w/o CL, which trains the policy from scratch, (2) 2-period CL, which trains the policy in obstacle-free environment, then in mixed scenarios, (3) 3-period CL w/ Dynamic, which, unlike 2-period CL, trains the policy in dynamic scenarios before moving to mixed scenarios, and (4) Ours, which trains the policy through obstacle-free, static-obstacle-only, and mixed scenarios, sequentially. 

Trained with the same amount of data, our curriculum achieves superior overall performance. Comparing the performance of 2-period CL with 3-period CL w/ Dynamic, we observe that training to avoid static obstacles enhances the ability to avoid dynamic obstacles. This is because, when trained exclusively in dynamic obstacle scenarios, drones tend to fly backwards for obstacle avoidance and therefore fail to meet the flight objective within the given time. However, with the presence of columns during training, the policy learns to alter its formation and decelerates less as balls appear.



\subsection{Real World Deployment} \label{sec:real-world-exp}

We deploy our RL policy and the planning-based baselines on Crazyflie 2.0. Due to the limited processing power of Crazyflie’s onboard processors, we execute the RL policy and planning baselines on a computer, and send commands to the drones via Crazyradio. We acquire the ground truth positions of the drones and obstacles through a motion capture system and estimate their linear and angular velocities by calculating the differences in position and rotation between consecutive frames. For the RL policy, CTBR commands are transmitted via radio at 50 Hz, while for planning baselines, trajectory waypoints are sent through position commands at 10 Hz. 
Since Swarm-RL generates thrust commands at 100Hz that exceeds the communication restraints imposed by the radio, we do not deploy it on the real drone.



\subsubsection{Sim2Real Transfer}
For better Sim2Real transfer, we add domain randomization and CTBR clip during training. More specifically, we randomize the initial position of each drones within 0.05m of the original formation, and randomize the initial rotation within $[-0.1\pi, 0.1\pi]$. Also, to avoid extreme actions, we clip the policy output, setting $\omega$ within $[-\pi/4, \pi/4]$ rad/s and $c$ within $[0.4, 0.9]$*max\_thrust. 


\subsubsection{Results and Analysis}

\begin{table}[H]
\centering
\begin{tabular}{cccc}
 &  Swarm-Formation & R-Mader & Ours \\
  \hline
 \hline
 Static (2 columns) & 0.5 & 0.6 & \textbf{0.7} \\
 Dynamic (1 ball) & 0.3 & \textbf{0.8} & 0.7  \\
 Mixed (2 columns \& 1 ball) & 0.3 & 0.5 & \textbf{0.8} \\
\end{tabular}
\caption{CFR of all methods in real world deployment.}
\label{table:real}
\end{table}

In real world deployment, we consider three scenarios, i.e.,  with 2 static columns, with one manually thrown ball, and with both the columns and the ball. 
We repeat each experiment ten times and report the CFR of each method in Table \ref{table:real}.
For full video demonstration, please refer to the supplementary materials or the website. 

The results further demonstrate the strong Sim2Real transfer capability of our method.
Our method can perform aggressive dodging strategy to safely navigate through all obstacles while maintaining formation, whereas the baseline methods often fail to find a viable path. Swarm-Formation halts when the dynamic obstacle drops before the drone; R-Mader struggles with cluttered static obstacles and fails to keep formation. 
Through the zero-shot Sim2Real deployment, we validate the efficacy of the action smoothing objective. 


\section{Conclusion and Future Work}


In this paper, we employ multi-agent reinforcement learning to tackle the challenging task of maintaining formation among multiple UAVs while avoiding both static and dynamic obstacles during directed flight. Our approach involves a two-stage training pipeline. In the first stage, we search for a balanced weight of all objectives in a simplified task setting. In the second stage, we apply curriculum learning to navigate the large exploration space, deriving a feasible policy for the more complex task. We introduce an attention-based observation encoder to effectively capture the features of the drones and the environments, ensuring the scalability of the policy. Our approach outperforms the state-of-the-art baselines in terms of both CFR and FM. The effectiveness of our method is further validated through real-world deployment.

Currently, our approach assumes perfect communication between drones and is limited to a structured environment without vision-based sensing. Future improvements could involve developing a vision-based policy to handle obstacles and making the system robust to communication delays.

\bibliographystyle{IEEEtran}
\bibliography{IEEEfull}

\begin{thebibliography}{10}
\providecommand{\url}[1]{#1}
\csname url@rmstyle\endcsname
\providecommand{\newblock}{\relax}
\providecommand{\bibinfo}[2]{#2}
\providecommand\BIBentrySTDinterwordspacing{\spaceskip=0pt\relax}
\providecommand\BIBentryALTinterwordstretchfactor{4}
\providecommand\BIBentryALTinterwordspacing{\spaceskip=\fontdimen2\font plus
\BIBentryALTinterwordstretchfactor\fontdimen3\font minus \fontdimen4\font\relax}
\providecommand\BIBforeignlanguage[2]{{%
\expandafter\ifx\csname l@#1\endcsname\relax
\typeout{** WARNING: IEEEtran.bst: No hyphenation pattern has been}%
\typeout{** loaded for the language `#1'. Using the pattern for}%
\typeout{** the default language instead.}%
\else
\language=\csname l@#1\endcsname
\fi
#2}}

\bibitem{liu2016multirobot}
Y.~Liu and G.~Nejat, ``Multirobot cooperative learning for semiautonomous control in urban search and rescue applications,'' \emph{Journal of Field Robotics}, vol.~33, no.~4, pp. 512--536, 2016.

\bibitem{rao2023temporal}
N.~Rao, S.~Sundaram, and P.~Jagtap, ``Temporal waypoint navigation of multi-uav payload system using barrier functions,'' in \emph{2023 European Control Conference (ECC)}.\hskip 1em plus 0.5em minus 0.4em\relax IEEE, 2023, pp. 1--6.

\bibitem{swarm1}
X.~Zhou, J.~Zhu, H.~Zhou, C.~Xu, and F.~Gao, ``Ego-swarm: A fully autonomous and decentralized quadrotor swarm system in cluttered environments,'' in \emph{2021 IEEE international conference on robotics and automation (ICRA)}.\hskip 1em plus 0.5em minus 0.4em\relax IEEE, 2021, pp. 4101--4107.

\bibitem{swarm2}
X.~Zhou, X.~Wen, Z.~Wang, Y.~Gao, H.~Li, Q.~Wang, T.~Yang, H.~Lu, Y.~Cao, C.~Xu, \emph{et~al.}, ``Swarm of micro flying robots in the wild,'' \emph{Science Robotics}, vol.~7, no.~66, p. eabm5954, 2022.

\bibitem{edg}
C.~Toumieh and D.~Floreano, ``High-speed motion planning for aerial swarms in unknown and cluttered environments,'' \emph{arXiv preprint arXiv:2402.19033}, 2024.

\bibitem{30uav}
G.~V{\'a}s{\'a}rhelyi, C.~Vir{\'a}gh, G.~Somorjai, T.~Nepusz, A.~E. Eiben, and T.~Vicsek, ``Optimized flocking of autonomous drones in confined environments,'' \emph{Science Robotics}, vol.~3, no.~20, p. eaat3536, 2018.

\bibitem{pso}
R.~J. Amala Arokia~Nathan, I.~Kurmi, and O.~Bimber, ``Drone swarm strategy for the detection and tracking of occluded targets in complex environments,'' \emph{Communications Engineering}, vol.~2, no.~1, p.~55, 2023.

\bibitem{mader}
J.~Tordesillas and J.~P. How, ``Mader: Trajectory planner in multiagent and dynamic environments,'' \emph{IEEE Transactions on Robotics}, vol.~38, no.~1, pp. 463--476, 2021.

\bibitem{kondo2023robust}
K.~Kondo, R.~Figueroa, J.~Rached, J.~Tordesillas, P.~C. Lusk, and J.~P. How, ``Robust mader: Decentralized multiagent trajectory planner robust to communication delay in dynamic environments,'' \emph{IEEE Robotics and Automation Letters}, 2023.

\bibitem{puma}
K.~Kondo, C.~T. Tewari, M.~B. Peterson, A.~Thomas, J.~Kinnari, A.~Tagliabue, and J.~P. How, ``Puma: Fully decentralized uncertainty-aware multiagent trajectory planner with real-time image segmentation-based frame alignment,'' \emph{arXiv preprint arXiv:2311.03655}, 2023.

\bibitem{dream}
B.~B. {\c{S}}enba{\c{s}}lar, ``Decentralized real-time trajectory planning for multi-robot navigation in cluttered environments,'' Ph.D. dissertation, University of Southern California, 2023.

\bibitem{mrnav}
B.~{\c{S}}enba{\c{s}}lar, P.~Luiz, W.~H{\"o}nig, and G.~S. Sukhatme, ``Mrnav: Multi-robot aware planning and control stack for collision and deadlock-free navigation in cluttered environments,'' \emph{arXiv preprint arXiv:2308.13499}, 2023.

\bibitem{batra2022decentralized}
S.~Batra, Z.~Huang, A.~Petrenko, T.~Kumar, A.~Molchanov, and G.~S. Sukhatme, ``Decentralized control of quadrotor swarms with end-to-end deep reinforcement learning,'' in \emph{Conference on Robot Learning}.\hskip 1em plus 0.5em minus 0.4em\relax PMLR, 2022, pp. 576--586.

\bibitem{han2019multi}
X.~Han, J.~Wang, Q.~Zhang, X.~Qin, and M.~Sun, ``Multi-uav automatic dynamic obstacle avoidance with experience-shared a2c,'' in \emph{2019 International Conference on Wireless and Mobile Computing, Networking and Communications (WiMob)}.\hskip 1em plus 0.5em minus 0.4em\relax IEEE, 2019, pp. 330--335.

\bibitem{formation1}
Z.~Lin, W.~Ding, G.~Yan, C.~Yu, and A.~Giua, ``Leader--follower formation via complex laplacian,'' \emph{Automatica}, vol.~49, no.~6, pp. 1900--1906, 2013.

\bibitem{quan2023robust}
L.~Quan, L.~Yin, T.~Zhang, M.~Wang, R.~Wang, S.~Zhong, X.~Zhou, Y.~Cao, C.~Xu, and F.~Gao, ``Robust and efficient trajectory planning for formation flight in dense environments,'' \emph{IEEE Transactions on Robotics}, 2023.

\bibitem{f-rl-1}
Z.~Liu, J.~Li, J.~Shen, X.~Wang, and P.~Chen, ``Leader--follower uavs formation control based on a deep q-network collaborative framework,'' \emph{Scientific Reports}, vol.~14, no.~1, p. 4674, 2024.

\bibitem{f-rl-2}
A.~Khan, E.~Tolstaya, A.~Ribeiro, and V.~Kumar, ``Graph policy gradients for large scale robot control,'' in \emph{Conference on robot learning}.\hskip 1em plus 0.5em minus 0.4em\relax PMLR, 2020, pp. 823--834.

\bibitem{f-rl-4}
T.~A. Karag{\"u}zel, V.~Retamal, and E.~Ferrante, ``Onboard controller design for nano uav swarm in operator-guided collective behaviors,'' in \emph{2023 IEEE International Conference on Robotics and Automation (ICRA)}.\hskip 1em plus 0.5em minus 0.4em\relax IEEE, 2023, pp. 3268--3274.

\bibitem{f-rl-5}
M.~Dawood, S.~Pan, N.~Dengler, S.~Zhou, A.~P. Schoellig, and M.~Bennewitz, ``Safe multi-agent reinforcement learning for formation control without individual reference targets,'' \emph{arXiv preprint arXiv:2312.12861}, 2023.

\bibitem{f-rl-6}
Y.~Yan, X.~Li, X.~Qiu, J.~Qiu, J.~Wang, Y.~Wang, and Y.~Shen, ``Relative distributed formation and obstacle avoidance with multi-agent reinforcement learning,'' in \emph{2022 International Conference on Robotics and Automation (ICRA)}.\hskip 1em plus 0.5em minus 0.4em\relax IEEE, 2022, pp. 1661--1667.

\bibitem{formation2}
M.~C. De~Gennaro and A.~Jadbabaie, ``Formation control for a cooperative multi-agent system using decentralized navigation functions,'' in \emph{2006 American Control Conference}.\hskip 1em plus 0.5em minus 0.4em\relax IEEE, 2006, pp. 6--pp.

\bibitem{f-rl-3}
J.~Wang, J.~Cao, M.~Stojmenovic, M.~Zhao, J.~Chen, and S.~Jiang, ``Pattern-rl: Multi-robot cooperative pattern formation via deep reinforcement learning,'' in \emph{2019 18th IEEE International Conference On Machine Learning And Applications (ICMLA)}.\hskip 1em plus 0.5em minus 0.4em\relax IEEE, 2019, pp. 210--215.

\bibitem{f3}
S.~Zhao, ``Affine formation maneuver control of multiagent systems,'' \emph{IEEE Transactions on Automatic Control}, vol.~63, no.~12, pp. 4140--4155, 2018.

\bibitem{f4}
Z.~Han, L.~Wang, and Z.~Lin, ``Local formation control strategies with undetermined and determined formation scales for co-leader vehicle networks,'' in \emph{52nd IEEE Conference on Decision and Control}.\hskip 1em plus 0.5em minus 0.4em\relax IEEE, 2013, pp. 7339--7344.

\bibitem{swarm-formation}
L.~Quan, L.~Yin, C.~Xu, and F.~Gao, ``Distributed swarm trajectory optimization for formation flight in dense environments,'' in \emph{2022 International Conference on Robotics and Automation (ICRA)}.\hskip 1em plus 0.5em minus 0.4em\relax IEEE, 2022, pp. 4979--4985.

\bibitem{f6}
A.~T. Nguyen, J.-W. Lee, T.~B. Nguyen, and S.~K. Hong, ``Collision-free formation control of multiple nano-quadrotors,'' \emph{arXiv preprint arXiv:2107.13203}, 2021.

\bibitem{f7}
Y.~Zhou, L.~Quan, C.~Xu, G.~Xu, and F.~Gao, ``Sparse-graph-enabled formation planning for large-scale aerial swarms,'' \emph{arXiv preprint arXiv:2403.17288}, 2024.

\bibitem{f8}
J.~Alonso-Mora, E.~Montijano, M.~Schwager, and D.~Rus, ``Distributed multi-robot formation control among obstacles: A geometric and optimization approach with consensus,'' in \emph{2016 IEEE international conference on robotics and automation (ICRA)}.\hskip 1em plus 0.5em minus 0.4em\relax IEEE, 2016, pp. 5356--5363.

\bibitem{panther}
J.~Tordesillas and J.~P. How, ``Panther: Perception-aware trajectory planner in dynamic environments,'' \emph{IEEE Access}, vol.~10, pp. 22\,662--22\,677, 2022.

\bibitem{deeppanther}
------, ``Deep-panther: Learning-based perception-aware trajectory planner in dynamic environments,'' \emph{IEEE Robotics and Automation Letters}, vol.~8, no.~3, pp. 1399--1406, 2023.

\bibitem{obs2}
D.~Wang, T.~Fan, T.~Han, and J.~Pan, ``A two-stage reinforcement learning approach for multi-uav collision avoidance under imperfect sensing,'' \emph{IEEE Robotics and Automation Letters}, vol.~5, no.~2, pp. 3098--3105, 2020.

\bibitem{obs3}
J.~Alonso-Mora, T.~Naegeli, R.~Siegwart, and P.~Beardsley, ``Collision avoidance for aerial vehicles in multi-agent scenarios,'' \emph{Autonomous Robots}, vol.~39, pp. 101--121, 2015.

\bibitem{huang2023collision}
Z.~Huang, Z.~Yang, R.~Krupani, B.~{\c{S}}enba{\c{s}}lar, S.~Batra, and G.~S. Sukhatme, ``Collision avoidance and navigation for a quadrotor swarm using end-to-end deep reinforcement learning,'' \emph{arXiv preprint arXiv:2309.13285}, 2023.

\bibitem{kaufmann2022benchmark}
E.~Kaufmann, L.~Bauersfeld, and D.~Scaramuzza, ``A benchmark comparison of learned control policies for agile quadrotor flight,'' in \emph{2022 International Conference on Robotics and Automation (ICRA)}.\hskip 1em plus 0.5em minus 0.4em\relax IEEE, 2022, pp. 10\,504--10\,510.

\bibitem{yu2103surprising}
C.~Yu, A.~Velu, E.~Vinitsky, Y.~Wang, A.~Bayen, and Y.~Wu, ``The surprising effectiveness of ppo in cooperative, multi-agent games. arxiv 2021,'' \emph{arXiv preprint arXiv:2103.01955}.

\bibitem{xu2024omnidrones}
B.~Xu, F.~Gao, C.~Yu, R.~Zhang, Y.~Wu, and Y.~Wang, ``Omnidrones: An efficient and flexible platform for reinforcement learning in drone control,'' \emph{IEEE Robotics and Automation Letters}, 2024.

\end{thebibliography}

\end{document}